\def\year{2021}\relax
\documentclass[letterpaper]{article} 
\usepackage{aaai21}  
\usepackage{times}  
\usepackage{helvet} 
\usepackage{courier}  
\usepackage[hyphens]{url}  
\usepackage{graphicx} 
\usepackage{natbib}  
\usepackage{caption} 
\usepackage{color}
\usepackage{wrapfig}
\newcommand\Tau{\mathcal{T}}
\usepackage{appendix}
\usepackage{subfig}
\usepackage{chngpage}
\usepackage{booktabs}
\usepackage[ruled,vlined]{algorithm2e}
\usepackage{tabularx}

\usepackage{rotating}

\graphicspath{ {./images/} }
\usepackage{amssymb,amsmath, amsthm}
\usepackage[switch]{lineno}
\begin{document}

\title{Stress Testing of Meta-learning Approaches for Few-shot Learning}

\author {
    Aroof Aimen\thanks{Equal Contribution},
    Sahil Sidheekh\footnotemark[1], 
    Vineet Madan\footnotemark[1], 
    Narayanan C. Krishnan \\
}
\affiliations {
    Indian Institute of Technology, Ropar \\
  (2018csz0001, 2017csb1104, 2017csb1119, ckn) @iitrpr.ac.in
}



\maketitle

\begin{abstract}
Meta-learning (ML) has emerged as a promising learning method under resource constraints such as few-shot learning. ML approaches typically propose a methodology to learn generalizable models. In this work-in-progress paper, we put the recent ML approaches to a stress test to discover their limitations. Precisely, we measure the performance of ML approaches for few-shot learning against increasing task complexity. Our results show a quick degradation in the performance of initialization strategies for ML (MAML, TAML, and MetaSGD), while surprisingly, approaches that use an optimization strategy (MetaLSTM) perform significantly better. We further demonstrate the effectiveness of an optimization strategy for ML (MetaLSTM++) trained in a MAML manner over a pure optimization strategy. Our experiments also show that the optimization strategies for ML achieve higher transferability from simple to complex tasks.
\end{abstract}

\section{Introduction}

Rapidly learning new skills from limited experience is a fundamental trait of human intelligence.  Replicating similar capabilities in deep neural networks is a challenging task. Meta-learning (ML) approaches have emerged as a promising direction in this context, facilitating learning from a limited amount of labeled training data (also referred to as the few-shot learning). They can be broadly classified into initialization and optimization strategies. Initialization strategies such as Model Agnostic Meta-learning (MAML) \cite{DBLP:conf/icml/FinnAL17}, Meta Stochastic Gradient Descent (MetaSGD)\cite{DBLP:journals/corr/LiZCL17}, Task Agnostic Meta-learning (TAML) \cite{ DBLP:conf/cvpr/JamalQ19} learn an optimal model initialization that swiftly adapts to new tasks with limited training data. Optimization strategies such as Learn2Learn \cite{andrychowicz2016learning}, MetaLSTM \cite{ DBLP:conf/iclr/RaviL17} learn parametric optimizers that accelerate the adaptation of a model to new tasks. Though the two approaches have the common objective of enforcing generalization to unseen tasks, the difference in their methodology presents a diverse set of merits and caveats.

Initialization methods learn an optimal prior on the model parameters through the experience gained across various tasks. The strategies define the optimal prior to be equally close to the individual training tasks' optimal parameters. This helps the model to quickly adapt to unseen tasks from the same distribution. MAML learns the optimal prior by assuming that a model quickly learns unseen tasks with sparse data if it is trained and tested under similar circumstances \cite{vinyals2016matching}. Literature suggests that the optimal prior learned by MAML may still be biased towards some tasks \cite{DBLP:conf/cvpr/JamalQ19}. TAML overcomes the bias by explicitly minimizing the inequality in the optimally initialized model performance across a batch of tasks. 

However, as the task complexity increases, finding an optimal prior becomes challenging. Thus, though gradient descent takes off from a good initialization (better than random), attaining good performance also depends on the model's ability to traverse the loss surface. A good initialization alone is insufficient. Our experiments confirm the degradation in the performance of these models as the complexity of the tasks increases. 

The optimization strategy MetaLSTM \cite{ DBLP:conf/iclr/RaviL17}, learns recurrent parametric optimizer capable of capturing both task-specific and agnostic knowledge. The learned optimizer mimics the gradient-based optimization of the base model. The parametric optimizer uses the current base model's loss and gradients to output the model parameters for the next iteration during the adaptation process.  Thus, the optimizer can be viewed as employing dynamic learning rates dependent on the model's parameters and task data, unlike MetaSGD that only considers the former dependence.

However, optimization strategies have the overhead of learning additional parameters and have limited scalability \cite{DBLP:conf/icml/FinnAL17,DBLP:journals/corr/LiZCL17}. Coordinate-wise sharing of the optimizer's parameters across the base model's parameters \cite{andrychowicz2016learning, DBLP:conf/iclr/RaviL17} reduces the learning overhead. 

Initialization approaches have achieved promising results on sparse data ML settings. However. their capability on increasingly complex tasks has not been studied. In this work-in-progress paper, we conduct a stress test on initialization and optimization ML strategies against increasing task complexity. We also combine the two approaches to learn a parametric meta-optimizer MetaLSTM++, a version of MetaLSTM trained in the MAML manner. We show that MetaLSTM++ achieves significantly better performance with fewer adaptation steps on simple and complex tasks. Further, motivated by human learning tactics - where experience gained from simple tasks helps to learn challenging tasks gradually- we also examine the transferability of these strategies from simple to complex tasks and vice versa.

\section{Methodology}
\subsection{Problem Formulation and Notations}
Given a principal dataset $\mathcal{D}$ and associated distribution of tasks $P(\Tau)$, ML techniques create partitions - meta-sets $\mathcal{D}^{meta-train}$, $\mathcal{D}^{meta-validation}$ and $\mathcal{D}^{meta-test}$ for training the model, tuning the hyperparameters and evaluating the performance. Each meta-set is a collection of mutually exclusive episodes of tasks drawn from the distribution $P(\Tau)$ and each task $\Tau_i$ is associated with a dataset $\mathcal{D}_i$ comprised of disjoint sets $\{\mathcal{D}_i^{tr} , \mathcal{D}_i^{test}\}$. Each task is an N-way K-shot learning problem. ML techniques aim to learn an accurate base model $f$ parameterized by $\theta$ for an unseen task $\Tau_i$ when fine-tuned with few examples $\mathcal{D}_i^{tr}$ of the task.

\subsubsection{MAML} uses a nested iterative process to learn the task-agnostic optimal initialization $\theta^*$. In the inner iterations representing the task adaptation steps, $\theta^*$ is separately fine-tuned for each meta-training task $\Tau_i$ using $\mathcal{D}^{tr}_i$ to obtain $\theta_i$ through gradient descent on the loss $\mathcal{L}^{tr}_i$. Specifically, $\theta_i$ is initialized as $\theta^*$ and updated using $\theta_i \leftarrow \theta_i - \alpha \nabla_{\theta_i}\mathcal{L}^{tr}_i(f_{\theta_i})$. In the outer loop, the meta optimization is performed over $\theta^*$ using the loss $\mathcal{L}^{test}_i$ computed with the task adapted model parameters $\theta_i$ on held-out dataset $\mathcal{D}^{test}_i$. Specifically, during meta-optimization $\theta^* \leftarrow \theta^*-\beta \nabla_{\theta^*} \sum_{\Tau_i \sim P(\Tau)} \mathcal{L}^{test}_i(f_{\theta_i})$.  

\subsubsection{MetaSGD}improves upon MAML by learning parameter specific learning rates $\alpha$ in addition to the optimal initialization in a similar nested iterative procedure. 
Meta-optimization is performed on $\theta^*$ and $\alpha$ in the outer loop using the loss $\mathcal{L}^{test}_i$ computed on held-out dataset $\mathcal{D}^{test}_i$. Specifically, during meta-optimization $(\theta^*,\alpha)\leftarrow (\theta^*,\alpha) - \beta \nabla_{(\theta^*, \alpha)} \sum_{\Tau_i \sim P(\Tau)} \mathcal{L}^{test}_i(f_{\theta_i})$. Learning dynamic learning rates for each parameter of a model makes MetaSGD faster and more generalizable than MAML. A single adaptation step is sufficient to adjust the model towards a new task. However the dependence of the learning rate only on model parameters limits the capability of MetaSGD.

\subsubsection{TAML}aims to reduce the bias of the optimal initialization, learned through MAML, towards any task by explicitly minimizing the inequality among the performances of a batch's tasks. TAML uses statistical measures like the Theil index, Atkinson index, Generalized entropy index, and Gini coefficient to estimate the inequality among tasks' performances. In this work, we use the Theil index owing to the average best results, defined as $\dfrac{1}{B}\sum_{i=1}^B \dfrac{\mathcal L^{test}_i}{\Bar{\mathcal L}^{test}} \ln \dfrac{\mathcal L^{test}_i}{\Bar{\mathcal L}^{test}}$ for measuring the inequality among the performances of the tasks in a batch, where $B$ is number of tasks in a batch, $\mathcal L^{test}_i$ is loss of task $\Tau_i$ on held-out dataset $\mathcal{D}^{test}_i$ and $\Bar{\mathcal L}^{test}$ is the average test loss of a batch of tasks. For few-shot learning, TAML proposes entropy minimization to preclude the bias of  $\theta^*$ towards any task $\Tau_i$ by maximizing the entropy of the labels predicted by $f_{\theta^*}$ and minimizing the entropy of the labels predicted by adapted model $f_{\theta_i}$. This is equivalent to a maximum entropy prior over $\theta^*$ such that the initialized model is not biased to any task.

\subsubsection{MetaLSTM}learns an optimizer $\mathcal{M}$ parametrized by $\phi$ to supervise the optimization process of the base model ($f_\theta$). The parametric optimizer is an LSTM, which is inherently capable of performing bi-level learning due to its architecture. During adaptation of $f_\theta$ on $\mathcal{D}^{tr}_i$, the parametric optimizer $\mathcal{M}$ takes meta information characterized by current loss $\mathcal{L}^{tr}_i$ and gradients $\nabla_{\theta_ {i}}(\mathcal{L}^{tr}_i)$ as input and outputs the next set of parameters $\theta_i$. Internally, the cell state of $\mathcal{M}$ corresponds to $\theta$, and the cell state update in $\mathcal{M}$ resembles a learned and controlled gradient update as the emphasis on the previous parameters and the current update is regulated by the learned forget and input gates respectively. While adapting $f_\theta$ to $\mathcal{D}^{tr}_i$, meta-information about the trajectory on the loss surface across the adaptation steps is captured in the hidden states of $\mathcal{M}$, representing the task-specific information.  During meta-optimization, $\phi$ is updated based on the loss $\mathcal L^{test}_i$ of task computed on held-out dataset $\mathcal{D}^{test}_i$ to garner the common knowledge across tasks.

\subsubsection{MetaLSTM++} 
\begin{figure}[t]
\centering
 \includegraphics[width =0.8\linewidth]{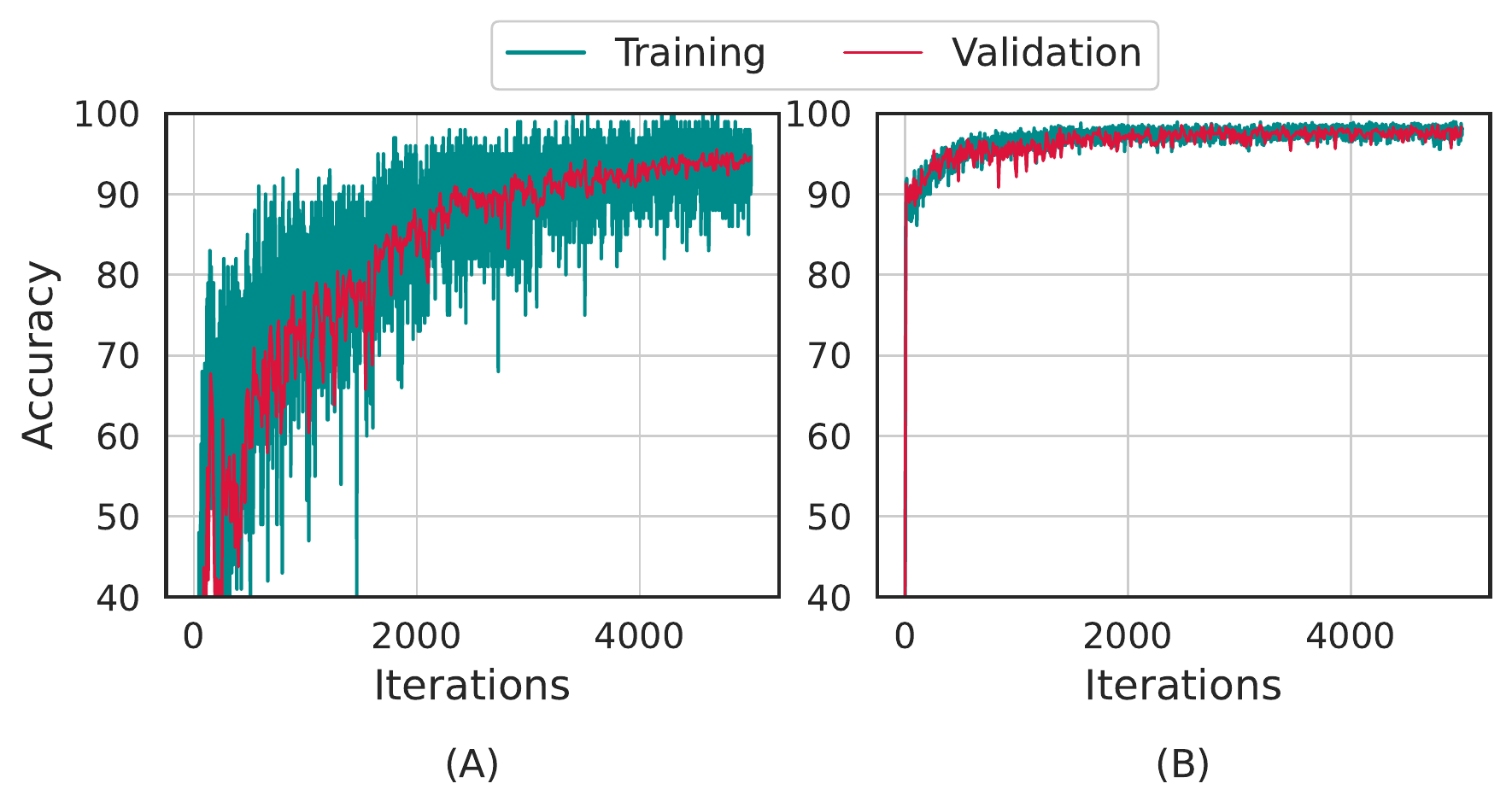}
  \caption{Oscillatory behaviour of MetaLSTM (A) as compared to MetaLSTM++ (B).}
\label{fig:oscillations}
\end{figure}

A caveat of MetaLSTM is the sequential update to the parametric optimizer $\mathcal{M}$ after each adaptation task. As a result, the optimization strategy traverses the loss surface in an ordered sequence of task specific optima. This leads to a longer and oscillatory optimization trajectory as shown in Figure \ref{fig:oscillations} and bias of $\mathcal{M}$ towards the final task. We propose to overcome this bottleneck by learning $\mathcal{M}$ according to the training procedure of initialization ML strategies, termed as MetaLSTM++. Unlike the MetaLSTM, the $\mathcal{M}$ of MetaLSTM++ is updated based on the average test loss of a task batch. This is intuitive as a batch of tasks may better approximate the data distribution, instead of a single task. The batch update on $\mathcal{M}$ makes the optimization trajectory smooth, short, and robust to task order (Figure \ref{fig:oscillations}). 

\section{Experiments and Results}
The ML approaches are benchmarked using the popular Omniglot dataset \cite{lake2015human}.  We consider this dataset owing to its simplicity and show that the performance of ML strategies on complex settings has not saturated even on this dataset. We follow the standard split (1200 : 423) of the dataset keeping 220 classes from the meta-training split to tune the models' hyperparameters. The images are downsampled to 28 $\times$28. We use the same architecture as in \cite{DBLP:conf/icml/FinnAL17} for the base model. We use a two-layer LSTM following \cite{DBLP:conf/iclr/RaviL17} for the parametric optimizer.  

The hyper-parameters for each ML approach have been fine-tuned for each complexity setting separately for a fair comparison. We find optimal hyper-parameters by performing a grid search over 30 different configurations for 5000 iterations. The search interval for all strategies is the same. The number of adaptation steps range within  $\left[2^2 - 2^6 \right]$, the meta and base learning rates follow log uniform distribution in the ranges $\left[1e^{-4} -  1 \right]$ and   $\left[1e^{-4} - 1e^{-2} \right]$ respectively. For TAML (Theil),   $\lambda$ also follows a log uniform distribution over the range of $\left[1e^{-2} - 1 \right]$. The meta batch size was varied among 4, 8, 16, and 32. The final model was trained for 20000 iterations using the optimal set of hyper-parameters. Early stopping was employed if no improvement was observed for 2500 steps. Cross-entropy is the loss function used in all the models.

We verified our implementation of all the ML approaches on the train-test split for the 20 way 1 and 5 shot settings. The results reported in Table \ref{20way1shot} are consistent with the findings in the existing literature. It is evident from the results that MetaLSTM++ outperforms the state of the art meta-learning approaches in both 1 and 5 shot settings, thus showing the promise of combining the initialization and the optimization strategies for ML.

\begin{table}{}
\centering
\label{accuracy_table}
\begin{tabular}{@{}lcc@{}}
\toprule 
 & \multicolumn{2}{c}{\textbf{Test Accuracy (20-Way)}} \\ 
\cmidrule(lr){2-3}
\multicolumn{1}{l}{\textbf{Model}} & 1 Shot & 5-Shot  \\
\cmidrule(lr){1-3}
MAML$^*$ & 91.93 $\pm$ 0.72 & 97.65 $\pm$ 0.20  \\ 
MetaSGD$^*$ & 94.58 $\pm$ 0.59 & 97.79 $\pm$ 0.23 \\ 
TAML  (Theil Index)$^*$ & 92.25 $\pm$ 0.70 & 95.14 $\pm$ 0.87 \\ \addlinespace  
MetaLSTM$^*$ & 90.63 $\pm$ 0.83 & 97.11 $\pm$ 0.24 \\ 
\textbf{MetaLSTM++} (Ours)  & \textbf{96.50 $\pm$ 0.42} & \textbf{98.41 $\pm$ 0.31} \\ \addlinespace
\bottomrule
\end{tabular}
\caption{Few shot classification performance of ML algorithms on the Omniglot dataset. The $\pm$ represents the 95\% confidence interval across 300 tasks. All the algorithms are rerun (denoted by *) on the same split for a fair comparison.}
\label{20way1shot}
\end{table}

\subsection{Increasing Task Complexity}
We use the most challenging few-shot learning setting, the one-shot setting, for the next set of experiments. Furthermore, increasingly challenging learning tasks are designed by increasing the number of classes.  In particular, we consider 20, 40, 60, 80, 100, 150, 175, and 200 classes. This challenging setting was never studied before.
\begin{figure}[t]
\centering
 \includegraphics[width =0.83\linewidth]{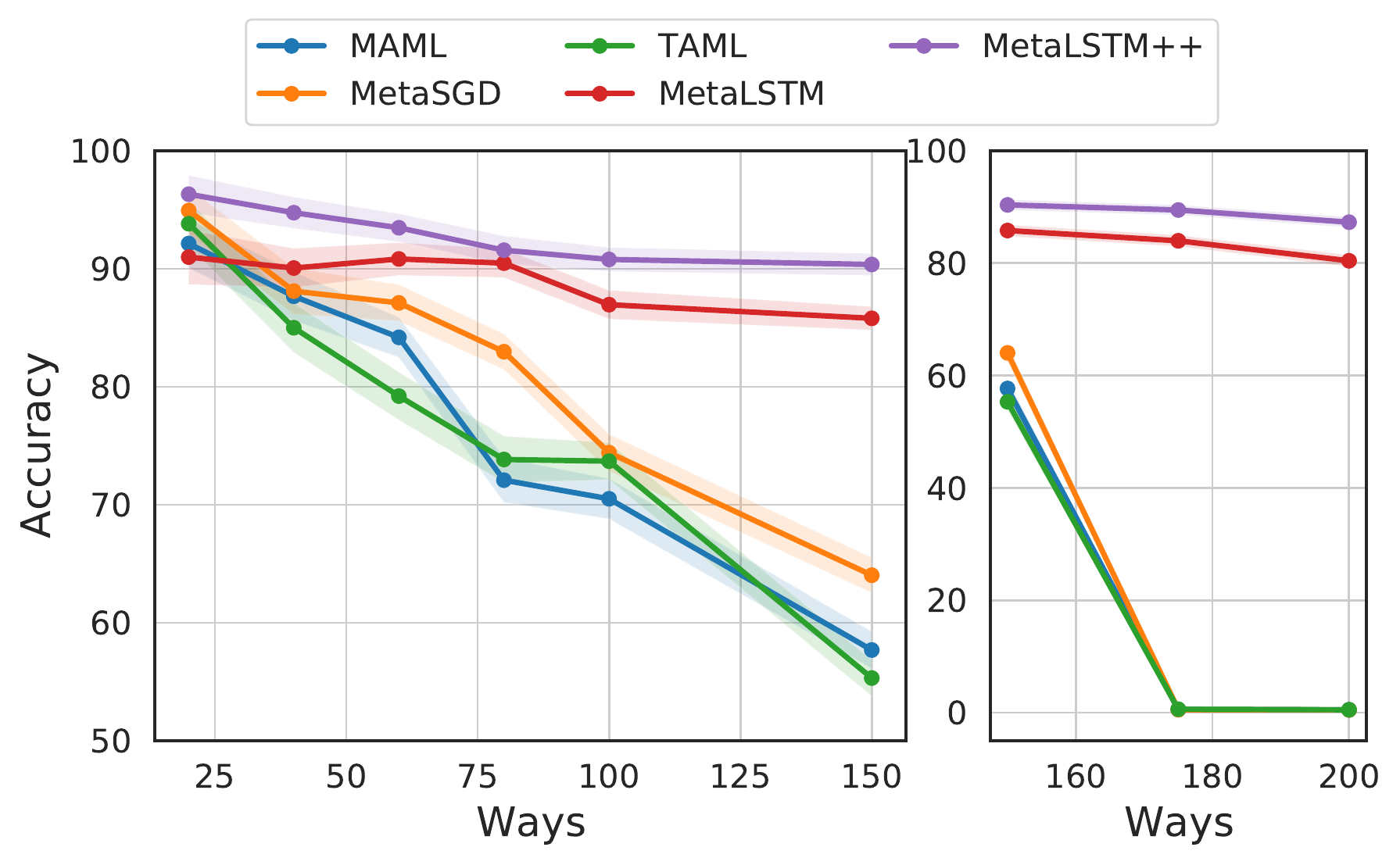}
  \caption{Accuracy of ML strategies w.r.t the increase in the number of ways in a task. The confidence interval is 99.9\%.}
\label{fig:accuracyWays}
\end{figure}
Figure \ref{fig:accuracyWays} shows the results for MAML, TAML, MetaSGD, MetaLSTM, and MetaLSTM++ as the number of ways in a task increases. The most salient observation is the rapid decrease in MAML, TAML, and MetaSGD performances against increasingly challenging learning tasks.  
 Further, an accelerated fall in their performance is witnessed when the number of ways in a task rises beyond 150. Despite the exhaustive hyperparameter search and finetuning, the initialization based approaches saturate at accuracies close to zero for 175 and 200 way tasks. This indicates that an optimal initialization alone is insufficient for the model to perform well on complex tasks. A justification for this behavior is that the optimal parameters for the diverse tasks may lie far from each other, so finding an initialization that lies in the proximity of all tasks may be difficult. Among initialization based methods, MetaSGD is comparatively robust to the increasingly complex tasks.  A possible explanation for this could be that MetaSGD learns parameter adaptive learning rates in addition to the optimal initialization, which helps the model traverse the loss surface to some extent while meta-testing.
 
MetaLSTM and MetaLSTM++, on the contrary, exhibit only a marginal decline in performance despite the increasing task complexity. This indicates that the parametric optimizers can learn the loss surface's generic dynamics to assist the model to approach the optima, even for complex tasks. It is also evident that MetaLSTM++ consistently outperforms all ML approaches, indicating that the parametric optimizer trained according to the initialization strategies has the edge over pure optimization and initialization strategies.

\subsection{Transferability across Complexity}
\begin{figure}[t]
\centering
 \includegraphics[width =\linewidth]{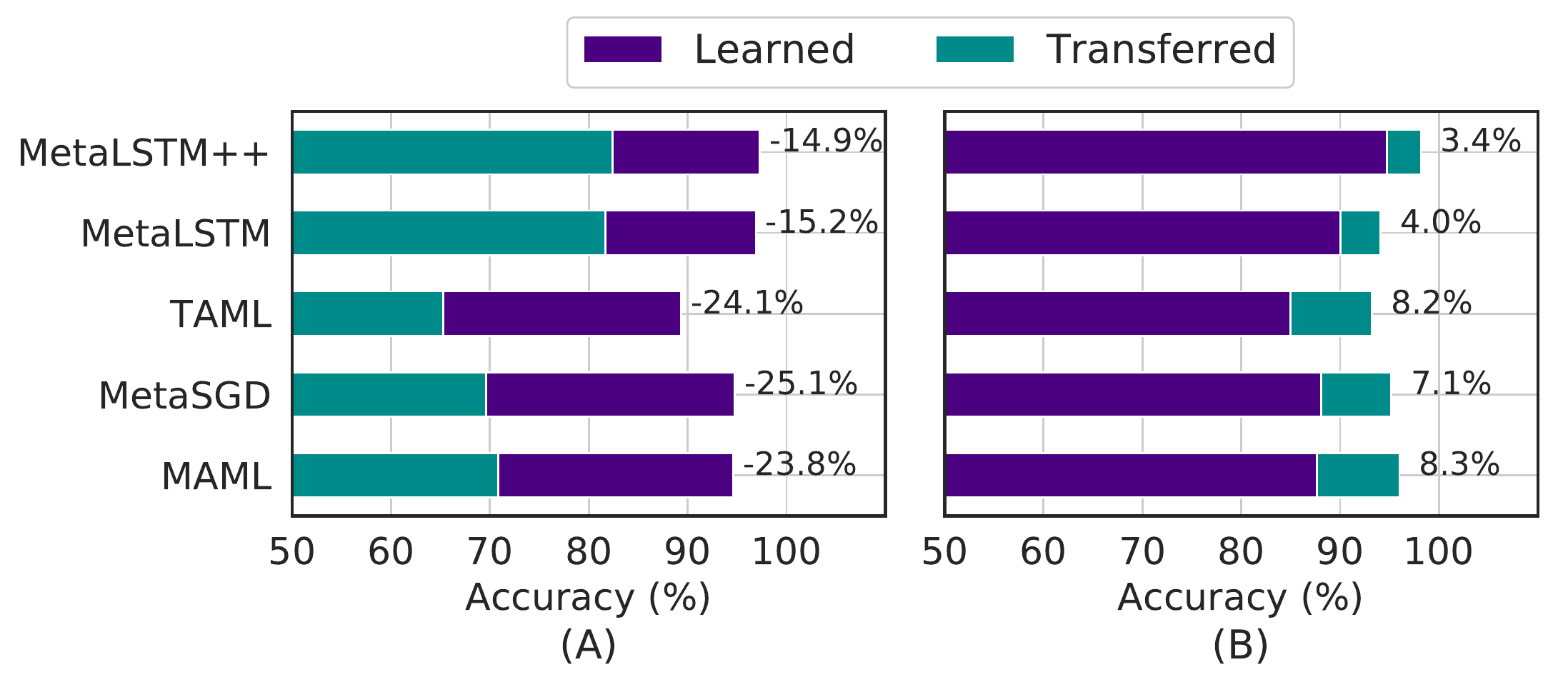}
  \caption{Transferability of ML approaches from (a) simple to complex tasks (40-way 5-shot to 40-way 1-shot) (b) complex to simple tasks (40-way 1-shot to 40-way 5-shot)}
\label{fig:transferibility}
\end{figure}

The primary goal of ML approaches is to learn from experience a prior that generalizes well. We investigate the generalizability of ML approaches on complex tasks, using prior learned from simpler tasks. Specifically,  we meta-train a model on a 40-way 5-shot setting and meta-test it on a 40-way 1-shot setting. A 5-shot learning task is less challenging than a 1-shot task. We observe a performance drop across all ML strategies as expected, as illustrated in Figure \ref{fig:transferibility}(A). However, the critical observation is that initialization strategies - MAML, MetaSGD, and TAML experience a substantial performance reduction, indicating that optimal initialization obtained for a model in a simple setting is not adequately generalizable to a complex setting. However, MetaLSTM and MetaLSTM++ show lesser performance reduction, confirming the generalizability of a parametric optimizer from simple to complex settings. MetaLSTM++ consistently outperforms other approaches.
We also investigate the reverse transferability of ML models from complex to simpler tasks.  In particular, we train and test a meta-model on a 40-way 1-shot and 5-shot tasks respectively. As the model sees more information during the meta-test time, we expect all the models' performance to increase. The results from Figure \ref{fig:transferibility}(B) show that the increase in the performance of MAML, TAML, and MetaSGD is higher than MetaLSTM and MetaLSTM++. However, MetaLSTM++ still achieves higher overall accuracy across both scenarios.

\begin{figure}[t]
\centering
\begin{tabular}{cc}
\subfloat[]{\includegraphics[width = 1.2in]{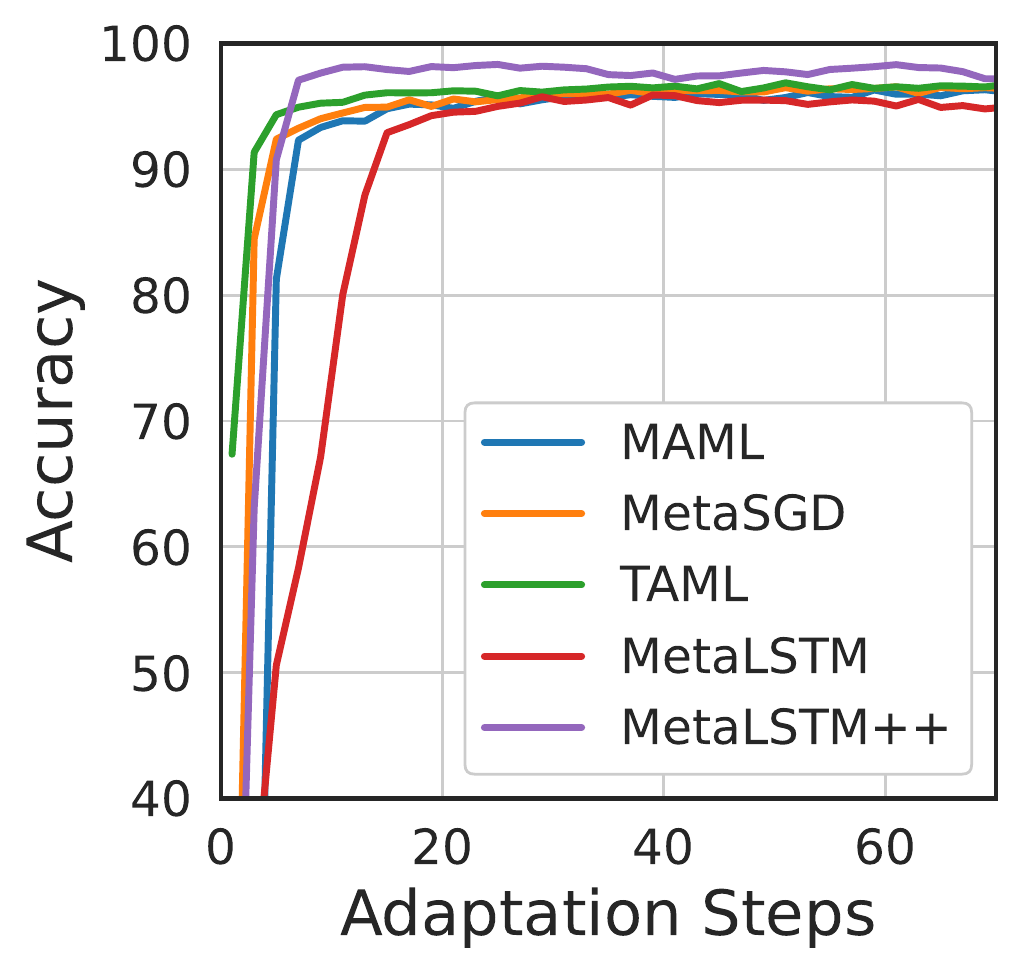}} &
\subfloat[]{\includegraphics[width = 2in]{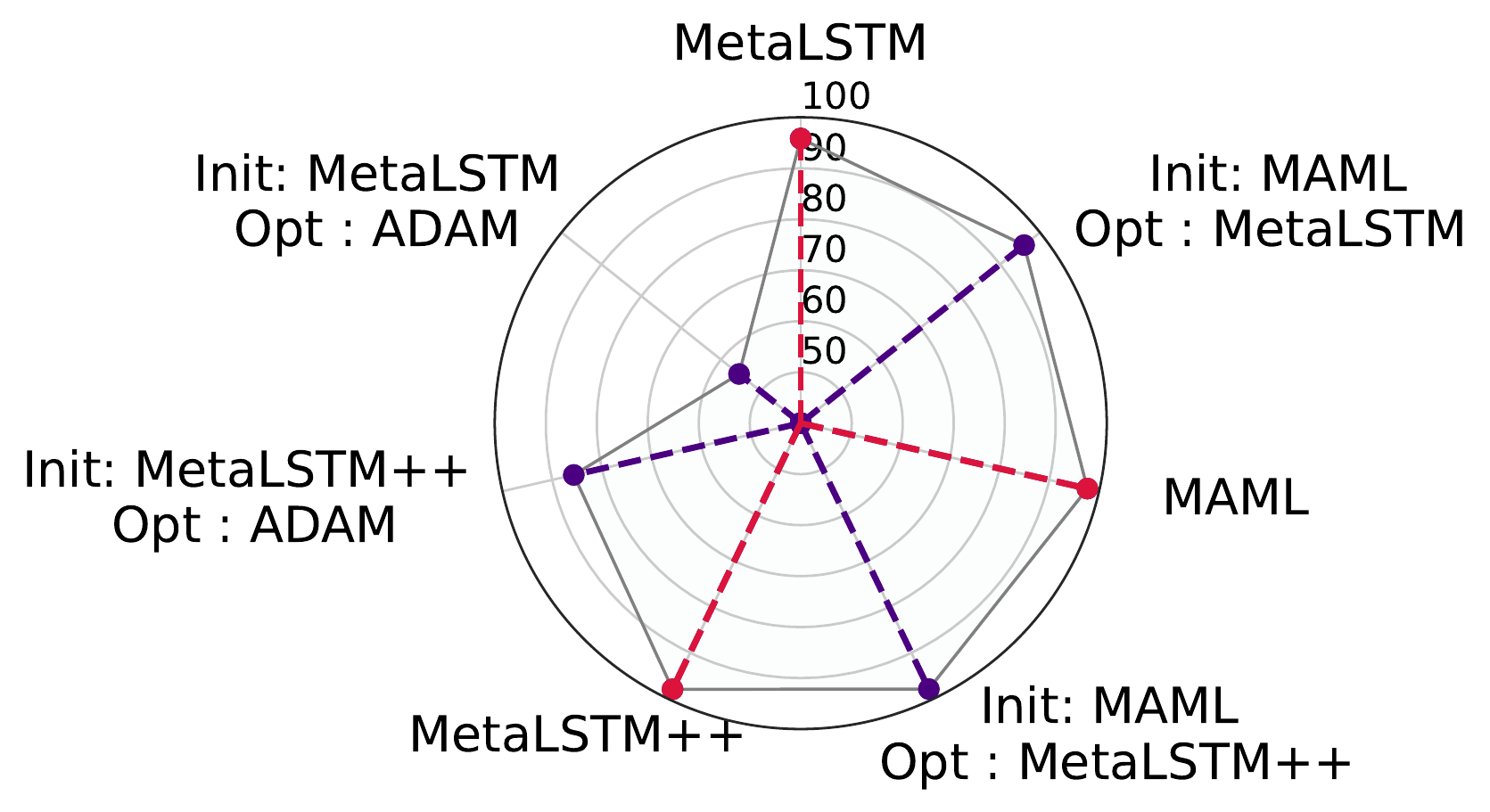}} 
\end{tabular}
\caption{Ablation studies across: (a) Adaptation steps (b) Initialization (radius = test accuracy across 300 tasks)}
\label{fig:ablation}
\end{figure}

To rule out the inadequacy of the adaptation steps during meta-testing, we observe the behavior of 40-way 1-shot models against an increasing number of adaptation steps during meta-testing on 40-way 5 shot tasks. The test accuracies are averaged across 300 tasks. We observe from Figure \ref{fig:ablation} (a) that initialization methods and MetaLSTM++ require less adaptation on the test data to achieve peak performance, but MetaLSTM requires significantly more adaptation steps. We also notice that MetaLSTM++ performs better than other meta-learning strategies throughout the scenario (achieves higher accuracy in a lesser number of adaptation steps).

\subsection{Ablation on the Initialization}
MetaLSTM and MetaLSTM++ also implicitly learn a good initialization for the base learner. We study the effect of decoupling the learned initialization from the parametric optimizer in a 20-way 5-shot setting. In the first experiment, we switch the MetaLSTM and MetaLSTM++ optimizers with Adam while retaining their initializations. This setting would be comparable to MAML.  We observe from Figure \ref{fig:ablation} (b) that a model initialized with MetaLSTM cell state and adapted with Adam performs significantly poorly at the test time, indicating that the initialization learned by MetaLSTM is substandard. We also observe that MetaLSTM++ learns a better initialization than MetaLSTM owing to its meta-training strategy; however, MAML learns the most desirable initialization. In the second experiment, we investigate the model's performance when initialized by MAML and guided by MetaLSTM and MetaLSTM++ optimizers. The results from Figure \ref{fig:ablation} (b) show no substantial increase in the performance, indicating that the learned parametric optimizers can guide the model to superior performance even with substandard initialization (MetaLSTM and MetaLSTM++).

\section{Conclusion}
Meta-learning approaches for few-shot learning have shown promising results on simpler tasks of the Omniglot benchmarking dataset. In this paper, we conduct a stress test on these approaches using more challenging one-shot learning tasks from the same dataset. We observe a sharp drop in the accuracy (close to 0!) for MAML, TAML, and MetaSGD with a significant increase in the number of classes (175). Surprisingly, optimization strategies like MetaLSTM and the proposed variant MetaLSTM++ continue to maintain accuracies above 80\% even with the challenging task of 200-way 1-shot learning. The experiments on transferability of meta-models from simpler to complex tasks (and vice versa) also suggest the effectiveness of optimization strategies over initialization.  While the results show the limitations of and motivate future research on pure initialization strategies for meta-learning, it also warrants the study into the causes for optimization strategies' effectiveness.

\section{Acknowledgements}
The support and the resources provided by ‘PARAM Shivay Facility’ under the National Supercomputing Mission, Government of India at the Indian Institute of Technology, Varanasi and under Google Tensorflow Research award are gratefully acknowledged.

\bibliography{bibfile}

\begin{thebibliography}{7}
\providecommand{\natexlab}[1]{#1}
\providecommand{\url}[1]{\texttt{#1}}
\providecommand{\urlprefix}{URL }
\expandafter\ifx\csname urlstyle\endcsname\relax
  \providecommand{\doi}[1]{doi:\discretionary{}{}{}#1}\else
  \providecommand{\doi}{doi:\discretionary{}{}{}\begingroup
  \urlstyle{rm}\Url}\fi

\bibitem[{Andrychowicz et~al.(2016)Andrychowicz, Denil, Gomez, Hoffman, Pfau,
  Schaul, Shillingford, and De~Freitas}]{andrychowicz2016learning}
Andrychowicz, M.; Denil, M.; Gomez, S.; Hoffman, M.~W.; Pfau, D.; Schaul, T.;
  Shillingford, B.; and De~Freitas, N. 2016.
\newblock Learning to learn by gradient descent by gradient descent.
\newblock In \emph{NIPS}, 3981--3989.

\bibitem[{Finn, Abbeel, and Levine(2017)}]{DBLP:conf/icml/FinnAL17}
Finn, C.; Abbeel, P.; and Levine, S. 2017.
\newblock Model-Agnostic Meta-Learning for Fast Adaptation of Deep Networks.
\newblock In \emph{{ICML}}, volume~70 of \emph{PMLR}, 1126--1135.

\bibitem[{Jamal and Qi(2019)}]{DBLP:conf/cvpr/JamalQ19}
Jamal, M.~A.; and Qi, G. 2019.
\newblock Task Agnostic Meta-Learning for Few-Shot Learning.
\newblock In \emph{{CVPR}}, 11719--11727.

\bibitem[{Lake, Salakhutdinov, and Tenenbaum(2015)}]{lake2015human}
Lake, B.~M.; Salakhutdinov, R.; and Tenenbaum, J.~B. 2015.
\newblock Human-level concept learning through probabilistic program induction.
\newblock \emph{Science} 350(6266): 1332--1338.

\bibitem[{Li et~al.(2017)Li, Zhou, Chen, and Li}]{DBLP:journals/corr/LiZCL17}
Li, Z.; Zhou, F.; Chen, F.; and Li, H. 2017.
\newblock Meta-SGD: Learning to Learn Quickly for Few-Shot Learning.

\bibitem[{Ravi and Larochelle(2017)}]{DBLP:conf/iclr/RaviL17}
Ravi, S.; and Larochelle, H. 2017.
\newblock Optimization as a Model for Few-Shot Learning.
\newblock In \emph{{ICLR}}.

\bibitem[{Vinyals et~al.(2016)Vinyals, Blundell, Lillicrap, Wierstra
  et~al.}]{vinyals2016matching}
Vinyals, O.; Blundell, C.; Lillicrap, T.; Wierstra, D.; et~al. 2016.
\newblock Matching networks for one shot learning.
\newblock In \emph{NIPS}, 3630--3638.

\end{thebibliography}
\end{document}